\documentclass{article}
\usepackage{graphicx}
\usepackage{booktabs}
\usepackage[numbers]{natbib}
 \usepackage[dblblindworkshop, final]{neurips_2025}

\workshoptitle{MATH-AI}

\usepackage[utf8]{inputenc} % allow utf-8 input
\usepackage[T1]{fontenc}    % use 8-bit T1 fonts
\usepackage{hyperref}       % hyperlinks
\usepackage{listings}
\usepackage{url}            % simple URL typesetting
\usepackage{booktabs}       % professional-quality tables
\usepackage{amsfonts}       % blackboard math symbols
\usepackage{nicefrac}       % compact symbols for 1/2, etc.
\usepackage{microtype}      % microtypography
\usepackage{xcolor}         % colors

\definecolor{codegreen}{rgb}{0,0.6,0}
\definecolor{codegray}{rgb}{0.5,0.5,0.5}
\definecolor{codepurple}{rgb}{0.58,0,0.82}
\definecolor{codered}{rgb}{0.8,0,0}

\lstdefinestyle{mystyle}{
    commentstyle=\color{codegreen},
    keywordstyle=\color{codered},
    numberstyle=\tiny\color{codegray},
    stringstyle=\color{codepurple},
    basicstyle=\ttfamily\footnotesize,
    breakatwhitespace=false,         
    breaklines=true,                 
    captionpos=b,                    
    keepspaces=true,                                    
    numbersep=5pt,                  
    showspaces=false,                
    showstringspaces=false,
    showtabs=false,                  
    tabsize=2
}

\title{LeanDojo-v2: A Comprehensive Library for AI-Assisted Theorem Proving in Lean}

\author{    
    Ryan Hsiang \\
  National Taiwan University\\
  b11901040@ntu.edu.tw \\
  \And William Adkisson \\
  Washington University in St. Louis\\
  adkisson@wustl.edu \\
  \And Robert Joseph George \\
  California Institute of Technology\\
  rgeorge@caltech.edu \\
  \And Anima Anandkumar \\
  California Institute of Technology\\
  anima@caltech.edu \\
}

\begin{document}

\maketitle

\begin{abstract}
Building on recent progress in machine learning for formal theorem proving in Lean, we introduce LeanDojo-v2: an end-to-end framework that unifies repository data extraction, LLM fine-tuning, proof search, AI autocompletion, and LLM-assisted theorem proving in a single software library. In addition, we provide Lean4Code, a lightweight IDE that integrates the various features of LeanDojo-v2 through a simple, mathematician-friendly UI. These tools will enable mathematicians and researchers to easily fine-tune language models on domain-specific Lean codebases and experiment with LLM-assisted theorem proving without a complex setup.
\end{abstract}

\section{Introduction}
Theorem proving with proof assistants such as Lean has enabled mathematicians to formalize mathematical concepts and proofs in a way that computers can verify
with absolute certainty. Lean serves as both a programming language and a proof assistant, enabling mathematicians to formalize complex mathematical structures and verify the correctness of their proofs mechanically. This formalization process ensures mathematical rigor and makes mathematical knowledge machine-readable and amenable to AI assistance. 

Machine learning, particularly with large language models (LLMs), has been shown to excel at solving mathematical problems. However, reasoning in natural language is unverifiable and prone to hallucinations and logical errors. Thus, combining machine learning with proof assistants such as Lean can make machine-generated proofs rigorous and verifiable. 

In recent years, various tools have been developed to integrate machine learning with the Lean theorem prover. LeanDojo\cite{yang2023leandojo} processes Lean repository information, such as premises and proofs, into training data. ReProver\cite{yang2023leandojo} trains retrieval-augmented language models using data generated by LeanDojo. LeanCopilot\cite{song2024lean} is an AI-assistant that integrates the ReProver models with the Lean 4 VSCode extension. LeanAgent\cite{kumarappan2025leanagent} extends upon ReProver by applying curriculum learning across different mathematical domains. For proof optimization, ImProver \cite{ahuja2024improveragentbasedautomatedproof} rewrite proofs that optimize for an arbitrary criterion. Other data-extraction frameworks have also been developed, such as the ntp-toolkit \cite{hu2025minictxneuraltheoremproving}, Lean4trace \cite{nesterov2024leantrace}, and LeanNavigator \cite{yin2025generatingmillionsleantheorems}.
 
Another essential tool for machine-based theorem proving in Lean is a read-eval-print loop (REPL), i.e., a programming interface that executes Lean tactics and returns the updated goal. Popular Lean REPLs include LeanDojo, Pantograph\cite{pantograph}, and Kimina Lean Server\cite{santos2025kiminaleanservertechnical}. In particular, Pantograph provides a Python interface for running Lean code, allowing users to apply a tactic to an unsolved goal and receive the resulting goal. 
% Kimina Lean Server is optimized for batch processing of Lean proofs, making it suitable for reinforcement learning with whole proof generation. 

Despite these efforts to advance the field of AI-assisted theorem proving, there are still significant technical hurdles in its implementation. Most of the tools described above are separate, research-oriented projects, which may be outdated and require nontrivial effort to maintain and update. In addition, all aforementioned tools require some level of programming/technical expertise to set up and use, which may be challenging for mathematicians without a programming background. 
% Codebases could benefit from further engineering work to meet production standards, more user-focused documentation, streamlined installation and deployment processes, and standardized testing practices.

% Additionally, the ReProver codebase, is primarily designed for decoder-only models and does not yet support modern encoder–decoder transformer architectures. This design choice has simplified the training pipeline but also means that fine-tuning newer models, such as DeepSeek \cite{ren2025deepseekproverv2advancingformalmathematical} and Kimina \cite{wang2025kiminaproverpreviewlargeformal}, is not directly supported.
To address these limitations, we introduce \textbf{LeanDojo-v2}, a unified infrastructure for Lean-based AI tools. It provides standardized APIs for data extraction, model training, and theorem proving, integrating frameworks such as LeanDojo and LeanAgent. Our contributions include an LLM fine-tuning pipeline, support for large-model inference, and theorem proving via both proof search and whole-proof generation.

In addition, we introduce \textbf{Lean4Code}, a lightweight Lean 4 native code editor. We build Lean4Code as a VSCodium-based code editor, leading to a familiar feel for current developers, and a user-friendly experience for mathematicians. Lean4Code also comes with the VSCode Lean4 extension already built in, making it the first Lean native code editor. Finally, Lean4Code comes with three custom extensions, an interactive user interface for LeanDojo, LeanCopilot, and ByteDance's Agent TARS\cite{qin2025ui}, providing an interactive interface for LeanDojo-v2 tools. 
%We present Lean4Code as the ultimate workspace for Lean4, containing the most comprehensive set of tools from State-of-the-art open source frameworks, with the most intuitive user experience for technical and non-technical users alike.

% Together, LeanDojo-v2 and Lean4Code bring the best capabilities of existing open-source frameworks into a single, unified ecosystem. By combining comprehensive functionality with an intuitive and accessible interface, our tools represent the most feature-rich and user-friendly environment for Lean theorem proving to date, powerful enough for advanced AI researchers, yet approachable for mathematicians and non-technical users alike.
\section{LeanDojo-v2}
\begin{figure}[h!]
    \centering
    \includegraphics[width=1\linewidth]{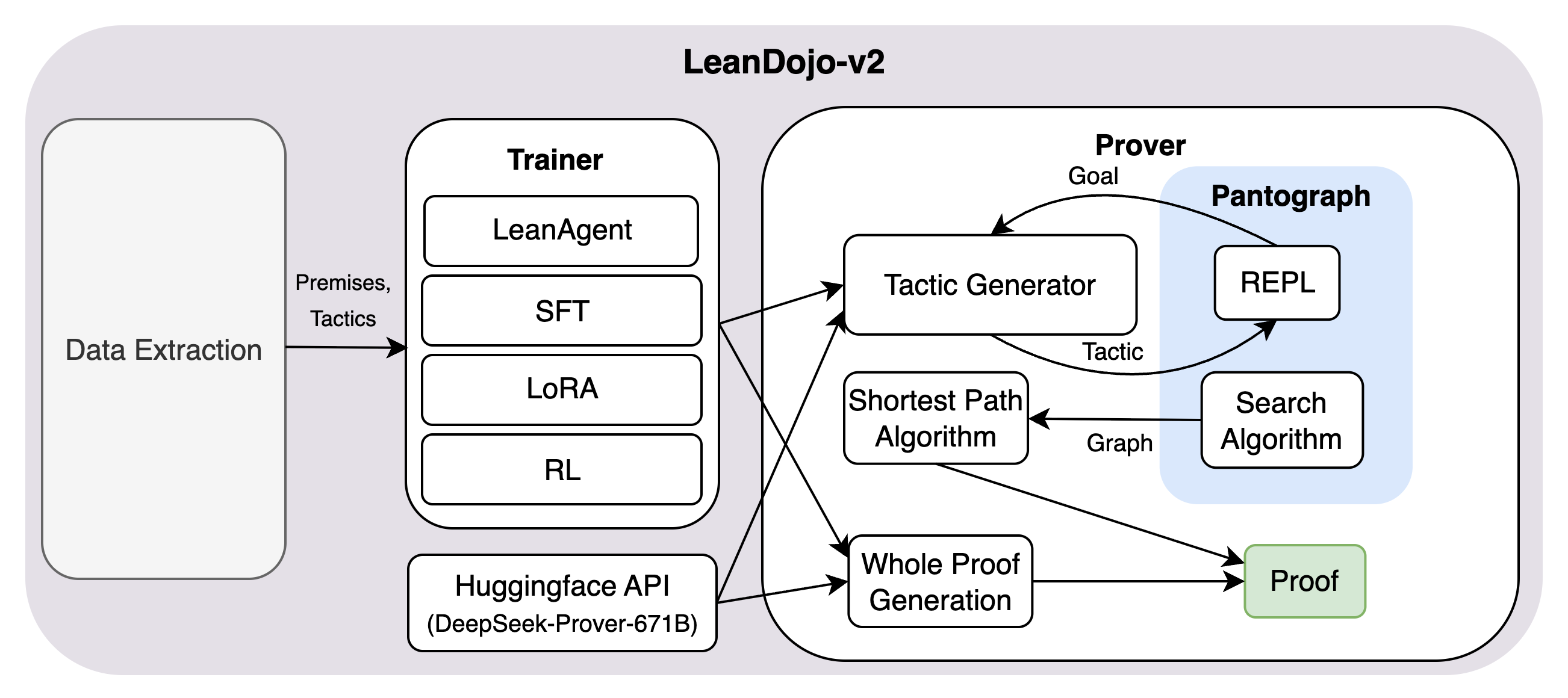}
    \caption{Overview of the LeanDojo-v2 architecture}
    \label{fig:LeanCopilotDemo}
\end{figure}
The primary goal of LeanDojo-v2 is to provide mathematicians and researchers with LLM-assisted theorem proving. However, existing pretrained LLMs remain insufficient for proving complex Lean theorems, as even the strongest models struggle with tactic generation on harder problems. To address this limitation, it is essential to fine-tune LLMs on the relevant data, enabling them to learn proof strategies specific to a given mathematical subfield and leverage data not captured by the original pretrained models. To this end, data collection and training on domain-specific Lean theorems and proofs are crucial.

LeanDojo-v2's training pipeline consists of three main components: data collection, training, and proving. For data collection, the user may select one or more Lean repositories by specifying their GitHub URL or local path. LeanDojo then extracts data such as abstract syntax trees, dependencies, theorems, proofs, and premises. This data is subsequently used to support two main methods of model training, through LeanAgent's trainer and with HuggingFace. 
\subsection{Training}
\paragraph{LeanAgent}
We categorize theorems extracted by LeanDojo into three difficulty levels—easy, medium, and hard—based on the 33rd and 67th percentiles of proof lengths, measured by the number of proof steps. The traced Lean repositories are then sorted according to the number of easy theorems they contain. Progressive training is performed on ReProver's premise retriever over these repositories, starting from the easiest and gradually moving to the harder ones.  
\paragraph{Fine-tuning HuggingFace models}
To support state-of-the-art LLM provers, we implement a flexible trainer for fine-tuning any HuggingFace model. Using the data collected by LeanDojo, we decompose the proofs of each theorem into a sequence of states and tactics. We use the proof state as input to the model and treat the associated tactic as the prediction target. Fine-tuning is supported through HuggingFace’s SFT trainer, with additional support for Low-Rank Adaptation (LoRA) \cite{hu2021loralowrankadaptationlarge} to reduce memory usage and training costs. We also implement reinforcement learning trainers, including GRPO \cite{shao2024deepseekmathpushinglimitsmathematical} and PPO \cite{schulman2017proximalpolicyoptimizationalgorithms}, and allow users to define custom reward functions such as syntax correctness and proof success. The training framework enables users to select any model and configuration or to extend the system by implementing their own training methods. For fine-tuning multiple Lean repositories, we use the same progressive approach as LeanAgent, sorting each repository by difficulty and performing one training run for each additional repository. 
\paragraph{LLMs through API inference}
However, the largest and most powerful LLMs, such as DeepSeek-Prover-V2-671B, are prohibitively large to fine-tune in practice. LeanDojo-v2 instead provides direct inference through HuggingFace API calls, allowing users to leverage their capabilities with a HuggingFace token and without the need for training. We integrate the HuggingFace API for AI autocompletion, enabling tactic suggestions to be displayed directly in Lean4Code through the Lean Infoview. Compared to LeanCopilot, this approach removes the need to install C++ inference libraries such as CTranslate2, reducing the build time from over ten minutes to just a few seconds. Moreover, the HuggingFace API eliminates the requirement to run model inference locally, enabling the use of more powerful hosted models while also reducing latency. 
\subsection{Theorem proving}
\begin{figure}[h!]
    \centering
    \includegraphics[width=0.9\linewidth]{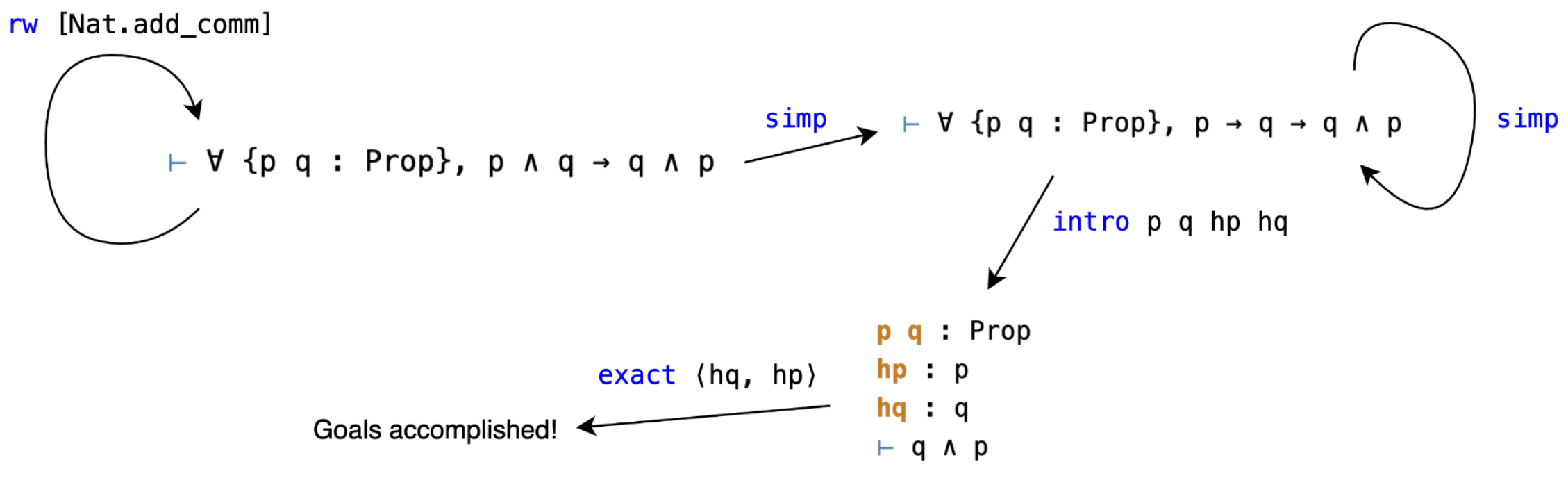}
    \caption{Illustration of a goal-tactic graph}
    \label{fig:LeanCopilotDemo}
\end{figure}
In Lean, a theorem is proved by applying a sequence of tactics, each of which updates the current goal representing the theorem’s state; when no goals remain, the proof is complete. For LLM-based theorem proving, there are two main approaches: step-by-step tactic generation and whole-proof generation.

Step-by-step tactic generation constructs proofs by sequentially generating tactics with an LLM tactic generator and updating the goal state through interaction with a REPL at each step. This process naturally formulates theorem proving as a search problem: each goal state corresponds to a node, each tactic corresponds to an edge, and the search begins from the initial node, i.e., the theorem’s starting goal state. By applying tactics, the prover traverses the graph until it reaches a terminal node with no remaining goals, at which point the theorem is proved. If a tactic fails or causes an error, the goal state remains unchanged. We use Pantograph as both the proof search algorithm and REPL. Pantograph supports both depth-first search (DFS) and Monte-Carlo tree search (MCTS), which is used to construct the goal-tactic graph. To recover the proof, we apply a shortest-path algorithm to reconstruct the tactic sequence that traverses from the initial goal state to the terminal state in the fewest steps, ensuring a minimal-length proof. We support LeanAgent, fine-tuned HuggingFace models, and API inference as tactic generators. The search agent itself is implemented as a base class, which can be extended to build custom provers with different tactic generators, including both machine learning–based and non-ML approaches. 

We also support whole-proof generation, where the complete proof of a theorem is produced in a single forward pass of the LLM. Users can either run inference locally on a trained HuggingFace model of their choice or use the HuggingFace API directly.

The library is designed to be modular and extensible. We implement trainers and provers for LeanAgent and HuggingFace, but we also provide abstract classes, e.g., \texttt{BaseAgent}, \texttt{BaseProver}, so that users can implement their own training and theorem-proving pipelines.

Below is an example usage of LeanDojo-v2 that extracts theorems and tactics from a Lean repository, performs supervised fine-tuning on the \texttt{DeepSeek-Prover-V2-7B} model, and conducts proof search to prove sorry theorems in the traced repository.
\begin{lstlisting}[language=Python]
from lean_dojo_v2.agent.hf_agent import HFAgent
from lean_dojo_v2.trainer.sft_trainer import SFTTrainer

url = "https://github.com/durant42040/lean4-example"
commit = "b14fef0ceca29a65bc3122bf730406b33c7effe5"

trainer = SFTTrainer(
    model_name="deepseek-ai/DeepSeek-Prover-V2-7B",
    output_dir="outputs-deepseek",
    epochs_per_repo=1,
    batch_size=2,
    lr=2e-5,
)

agent = HFAgent(trainer=trainer)

agent.setup_github_repository(url=url, commit=commit)

agent.train()
agent.prove()
\end{lstlisting}
We provide more examples of LeanDojo-v2 in Appendix \ref{appendix:code}. 
\section{Lean4Code}
While our LeanDojo-v2 tools and Lean itself are quite accessible to mathematicians, the configuration needed to get started with Lean is cumbersome, and more technical than many mathematicians may feel comfortable with. To combat this, we introduce Lean4Code, the first-ever Lean native code editor.
\subsection{VSCode Functionalities}
The first primary focus of Lean4Code is enabling frictionless interaction with Lean for all users, regardless of their technical background. With this in mind, we developed Lean4Code off of VSCodium, an open source version of VSCode that disables all VSCode default telemetry settings. We chose VSCodium for two reasons. Firstly, it's a familiar environment for developers. %Today's most popular code editors (VSCode, VSCodium, Cursor) are all based on VSCode. %By using VSCodium as the base of Lean4Code, users of VSCode-based code editors should have a frictionless experience using Lean4Code. 
Secondly, it has full access to the VSCode extension market, meaning any VSCode extension can be used within Lean4Code. This means users coming from VSCode retain all their extensions, while all users can access the full capabilities of the VSCode marketplace. %By using VSCodium as the base, Lean4Code has full access to the VSCode extension market. %Moreover, the nature in which VSCode extensions are created and used allows for a seamless integration of custom LeanDojo-v2-based extensions.
Most importantly, by building off of VSCodium, we are able to use the VSCode Lean4 extension. By pre-downloading this extension into our app, we have created the first-ever Lean4 native code editor.

\subsection{Extensions}
The second primary focus of Lean4Code is creating an interactive user interface for the LeanDojo-v2 project. To accomplish this, we modified the VSCode Lean4 extension and created three extensions of our own. 
\paragraph{Lean 4}
The first and most important extension inside Lean4Code is the VSCode Lean4 extension. A key feature of the VSCode Lean4 extension is automatic creation and configuration of a Lean4 project and it's environment. %The extension adds a "for all" symbol to the top right-hand side of the code editor, identifiable by an upside-down "A" symbol. 
%Upon clicking it, the user is presented with an option to automatically create the necessary files for a Lean project, which includes a Lean-toolchain file (specifying which version of Lean the project must use), a lakefile.toml or lakefile.lean file (a configuration file), a lake-maifest.json file (manifest for root package), and a .lake/directory file (temporary storage for .lake files). 
We modify this process by adding LeanDojo-v2 as a configuration folder, giving any Lean project created with the Lean4 extension access to the family of LeanDojo-v2 tools locally.

\paragraph{Custom Extensions}
Lean4Code also comes with three more custom extension panels representing the LeanDojo tools. Right now, Lean4Code has custom extension panels for LeanDojo, LeanDojo-v2, and Agent-TARS (ByteDance's agentic AI assistant). Each of these extensions is a true "one-click" experience, meaning they simply prompt users for the necessary input variables and take care of everything else behind the scenes.

\begin{table}[h!]
\centering
\begin{tabular}{ll}
\toprule
\textbf{Extension} & \textbf{Purpose} \\
\midrule
Lean4 (modified) & Downloads and configures Lean4/LeanDojo-v2 \\
LeanDojo & Supports one-click LeanDojo tracing \\
LeanCopilot & Calls models to complete theorems \\
Agent-TARS & General agentic AI assistance \\
\bottomrule
\end{tabular}
\caption{Lean4Code Extensions}
\label{tab:lean4code}
\end{table}

\paragraph{Limitations and Future Integration
} We acknowledge that Lean4Code has limitations. One potential limitation is changes to Lean versions, as newer Lean versions are often not backward compatible. This is not a concern, though, as we will be updating Lean4Code's built-in Lean4 extension with every major release. Moreover, Lean4Code currently calls on the LeanDojo-v2 tools by cloning LeanDojo-v2 into the desired project, meaning it supports the Lean4 version of the latest LeanDojo-v2 release.
We turn over Lean4Code to community development and welcome public contributions to ensure it best reflects the community's needs.

\section{Conclusion}
In this work, we introduce LeanDojo-v2 and Lean4Code, a unified framework for Lean+AI. LeanDojo-v2 supports supervised, parameter-efficient, and reinforcement learning based training methods. The user will be able to access AI-assisted theorem proving while only writing several lines of code, or through the graphical user interface of Lean4Code. We hope this framework will accelerate progress toward more capable and accessible AI-assisted theorem proving, ultimately supporting mathematicians in tackling increasingly sophisticated research problems.

Future work includes adding support for other machine learning methods and proof search algorithms, integrating LeanProgress, a theorem-proving framework that incorporates proof progress prediction, into LeanDojo-v2, and the integration of more theorem-proving tools into Lean4Code.  

\section*{Acknowledgments}
Ryan Hsiang is supported by the Summer Undergraduate Research Fellowships (SURF) program at Caltechiner for fine-tuning any HuggingFace model. Using the data collected by LeanDojo, we decompose the proofs of each theorem into a sequence of states and tactics. We use the proof state as input to the model and treat the associated tactic as the . Anima Anandkumar is supported by the Bren Named Chair, Schmidt AI 2050 Senior Fellowship, and ONR (MURI grant N00014-18-12624). 

\bibliographystyle{unsrtnat}
\bibliography{main}

\appendix
\newpage
\section{Lean4Code Implementation}
Lean4Code incorporates LeanDojo-v2 through the built-in extension panels. These extensions are custom to Lean4Code and are currently not available anywhere on the VSCode extension market.
\paragraph{LeanDojo}The LeanDojo extension panel prompts the user for five inputs: A project name, Lean repository GitHub URL, a commit hash for the target GitHub repository, a user's GitHub personal access token, and the Lean version of the project. The user then toggles their trace's build dependencies, determining whether or not to build the repository's dependencies, and then clicks "Create Project". Upon creation, the LeanDojo extension will automatically create a folder on the user's desktop with the project's name, before cloning the repository, LeanDojo-v2, and creating a Python script, a temp folder, and an output folder for tracing inside the project. 
From there, the user simply clicks one button to trace their repository, and a Python script with their specific repository details configured will run, dumping the trace artifacts into the "out" folder when the trace is complete.
\begin{figure}[h!]
    \centering
    \includegraphics[width=1\linewidth]{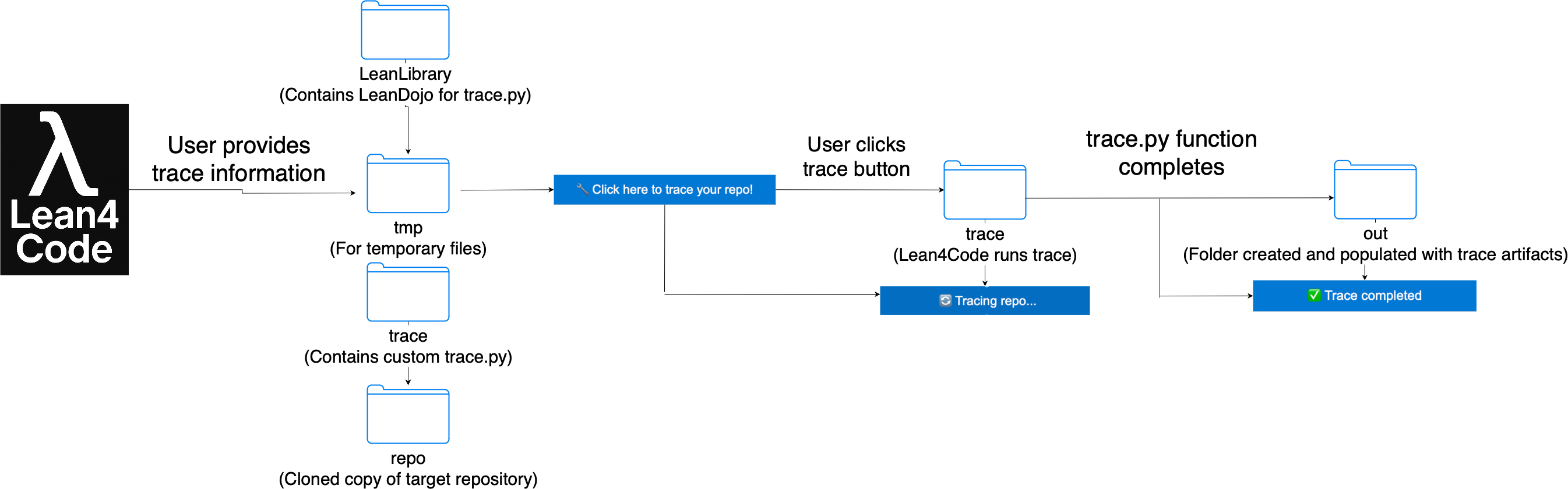}
    \caption{Overview of using LeanDojo in Lean4Code}
    \label{fig:LeanDojoFlowchart}
\end{figure}
\begin{figure}[h!]
    \centering
    \includegraphics[width=1\linewidth]{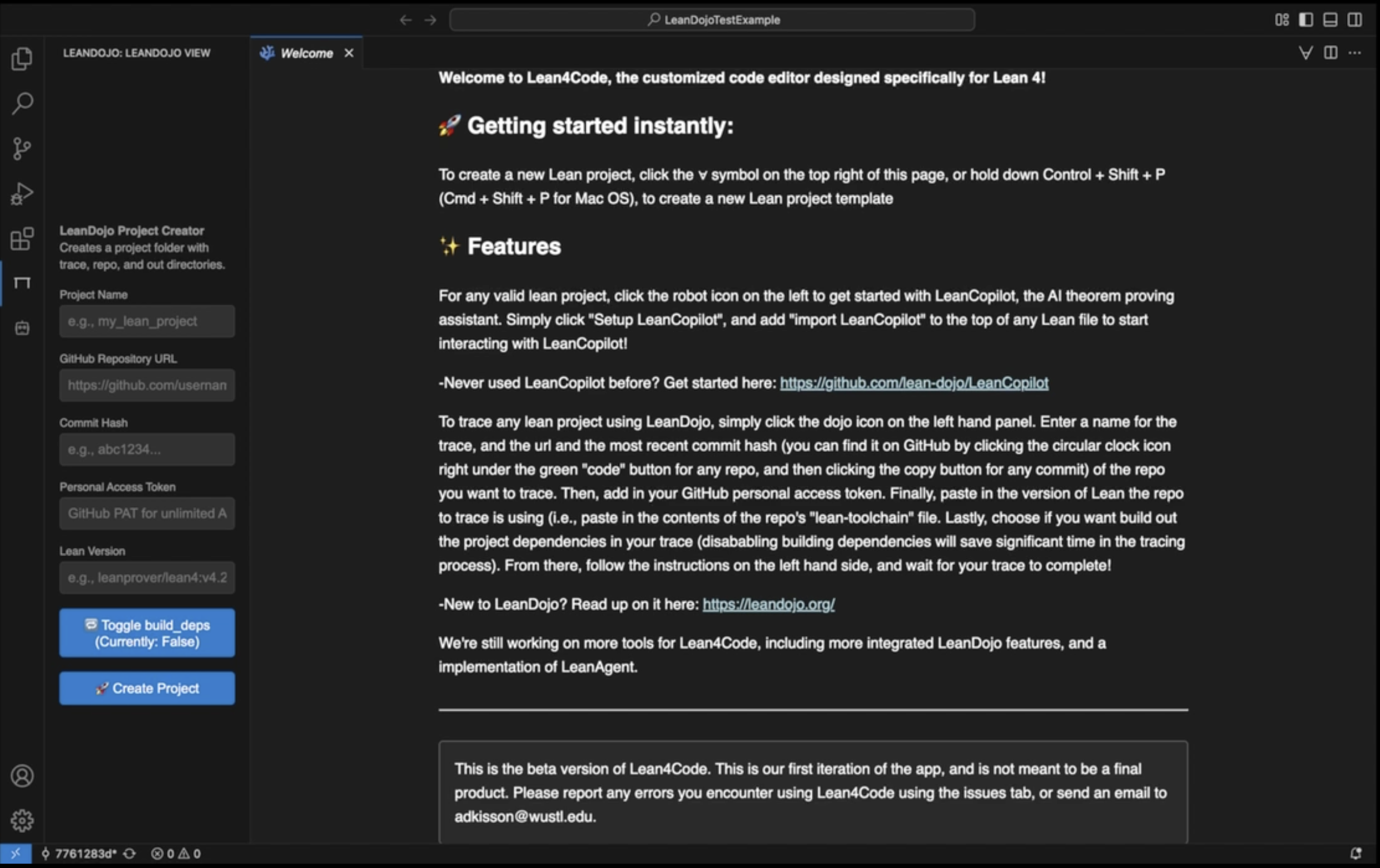}
    \caption{Input paramater screen for LeanDojo panel}
    \label{fig:LeanDojoScreenshot}
\end{figure}
\paragraph{LeanCopilot} While working on Lean projects, users may utilize the LeanCopilot extension panel for theorem proving assistance. The LeanCopilot extension panel allows users two options to integrate LeanCopilot into their project. Users may either download the Byte5 retriever model locally or make an API call to a HuggingFace model. If the user chooses the first option, we simply download a simple theorem-proving model locally and use said model to power LeanCopilot. If the user chooses the second option, they enter a HuggingFace personal access token, which calls the DeepSeek-Prover-V2-671B from HuggingFace's default inference provider. Then, we call the said model using a natural language prompt any time the user requests tactic suggestions.
After the user chooses their preferred method, the LeanCopilot extension automatically executes the LeanCopilot installation steps, including editing the project configuration file and installing the needed dependencies. LeanCopilot can help users complete proofs, mostly by suggesting the next tactic a user may need to complete a theorem. Given an incomplete theorem, calling "suggest tactics" will prompt LeanCopilot to suggest the next tactic for completing a theorem.
\begin{figure}[h!]
    \centering
    \includegraphics[width=1\linewidth]{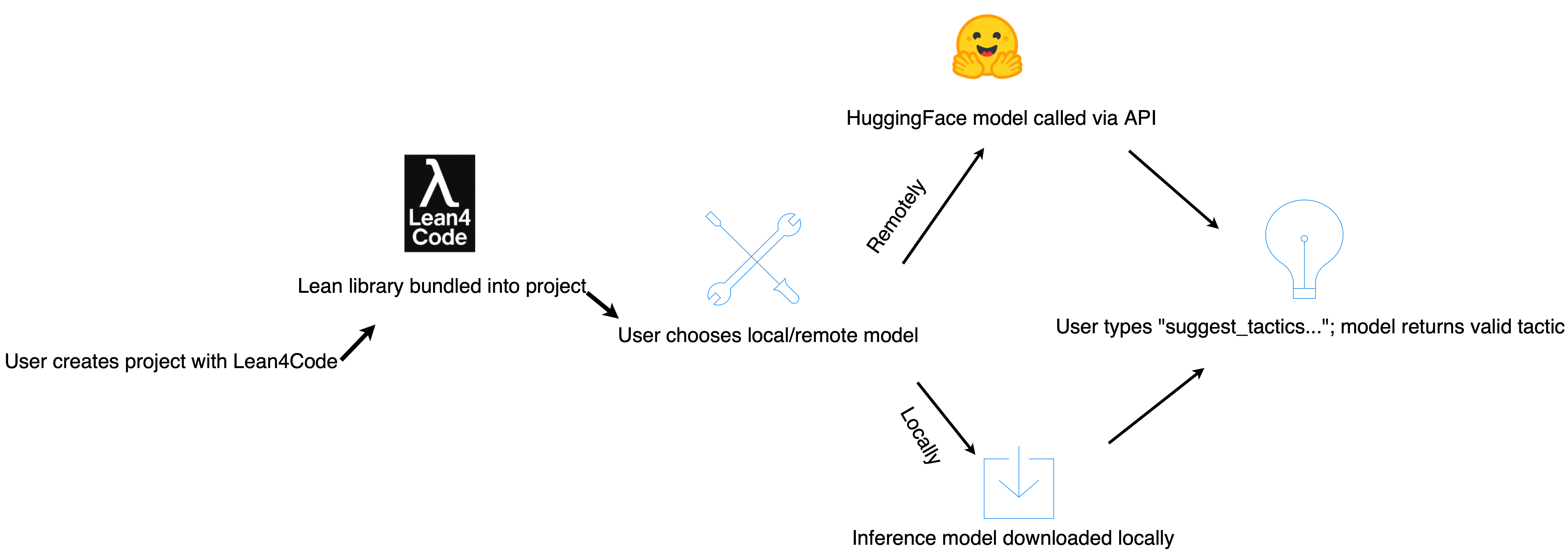}
    \caption{Overview of using LeanCopilot in Lean4Code}
    \label{fig:LeanCopilotFlowchart}
\end{figure}
\begin{figure}[h!]
    \centering
    \includegraphics[width=1\linewidth]{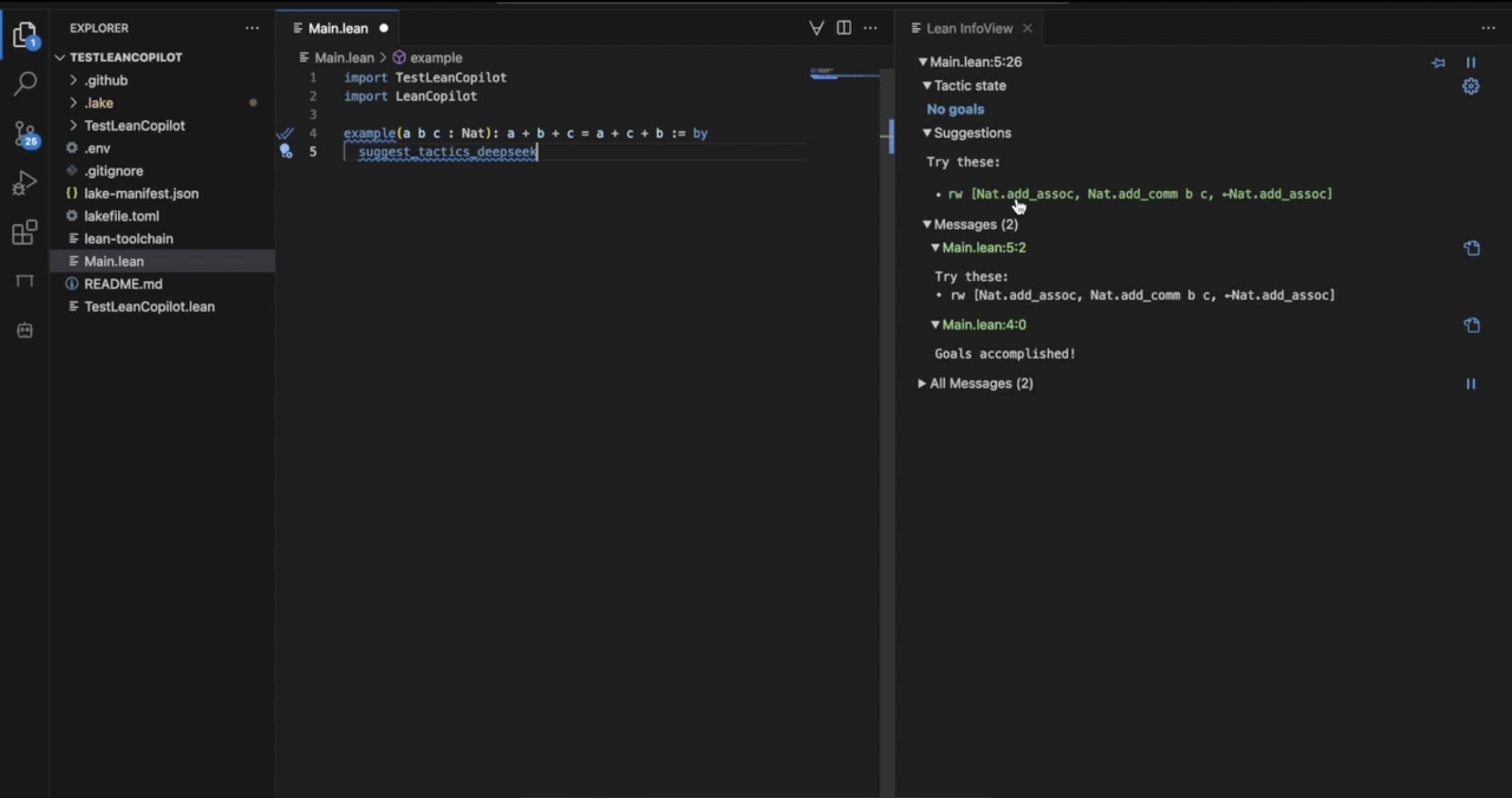}
    \caption{Prompting LeanCopilot for tactic suggestions}
    \label{fig:LeanCopilotScreenshot}
\end{figure}
\paragraph{TARS} The third and final extension we introduce in Lean4Code is the TARS extension panel. This extension provides an interactive user interface for any user to interact with ByteDance's Agent TARS, a multi-modal agentic AI. Upon opening the TARS extension, users are prompted for their OpenAI access token. From there, users may boot one of two versions of Agent TARS, either the Chat Only Mode or the Agentic Mode. If the user chooses to run Agent TARS in the Chat Only mode, Lean4Code will automatically call Agent TARS in the user's terminal, disabling browser control. If the user chooses to run Agent TARS in Agentic Mode, Lean4Code will follow the same process, except enabling Agent TARS's browser control, thinking, and planning functionality. 
\begin{figure}[h!]
    \centering
    \includegraphics[width=1\linewidth]{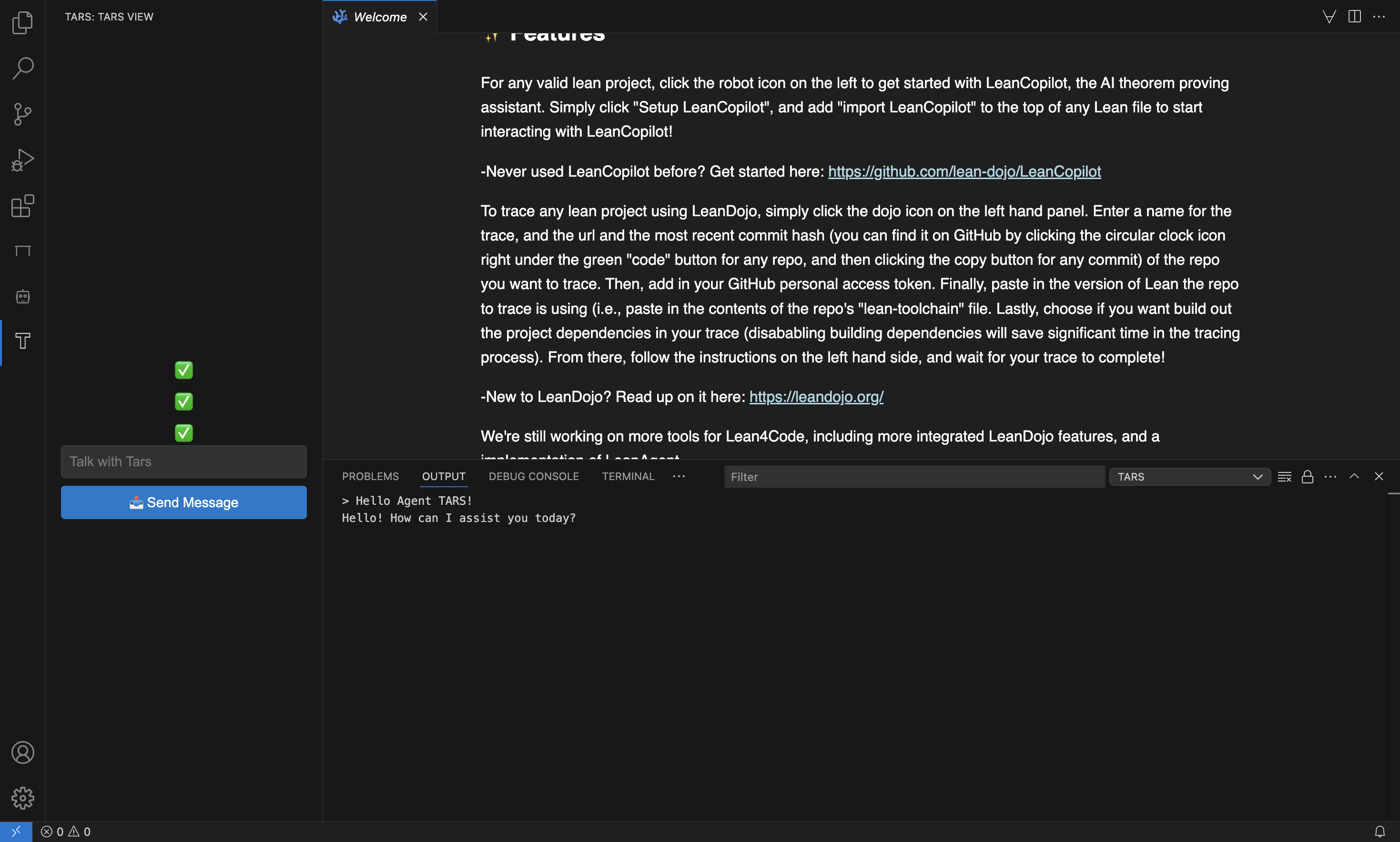}
    \caption{Chatting with Agent-TARS}
    \label{fig:TARSScreenshot}
\end{figure}

Once Agent TARS has been booted through the extension, Lean4Code provides the user with a space to send messages to their instance of Agent TARS. Upon receiving the first message, Lean4Code automatically creates another terminal instance and starts a new session there. Lean4Code will then store the session ID, which enables Agent TARS to use the user's previous messages from the chat as context. Both the user's messages and Agent TARS's response will appear in the new terminal window.
\newpage
\section{LeanDojo-v2 examples}\label{appendix:code}
In this section, we provide some example usages of LeanDojo-v2 for model training and automated theorem proving. 
% \subsection*{Example 1: Supervised fine-tuning}
% The following code extracts theorems and tactics from a Lean repository, performs supervised fine-tuning on the \texttt{DeepSeek-Prover-V2-7B} model, and conducts proof search to prove sorry theorems in the traced repository.
% \begin{lstlisting}[language=Python]
% from lean_dojo_v2.agent.hf_agent import HFAgent
% from lean_dojo_v2.trainer.sft_trainer import SFTTrainer

% url = "https://github.com/durant42040/lean4-example"
% commit = "b14fef0ceca29a65bc3122bf730406b33c7effe5"

% trainer = SFTTrainer(
%     model_name="deepseek-ai/DeepSeek-Prover-V2-7B",
%     output_dir="outputs-deepseek",
%     epochs_per_repo=1,
%     batch_size=2,
%     lr=2e-5,
% )

% agent = HFAgent(trainer=trainer)

% agent.setup_github_repository(url=url, commit=commit)

% agent.train()
% agent.prove()
% \end{lstlisting}

\subsection*{Example 1: Progressive training with LeanAgent}
The following code extracts theorems and tactics from two Lean repositories, performs progressive training with LeanAgent, and conducts proof search to prove sorry theorems in the traced repository.
\begin{lstlisting}[language=Python]
from lean_dojo_v2.agent.lean_agent import LeanAgent

url1 = "https://github.com/durant42040/lean4-example"
commit1 = "b14fef0ceca29a65bc3122bf730406b33c7effe5"

url2 = "https://github.com/durant42040/lean4-example"
commit2 = "005de00d03f1aaa32cb2923d5e3cbaf0b954a192"

agent = LeanAgent()

agent.setup_github_repository(url=url1, commit=commit1)
agent.setup_github_repository(url=url2, commit=commit2)

agent.train()
agent.prove()
\end{lstlisting}

\subsection*{Example 2: Fine-tuning with GRPO}
The following code extracts theorems and tactics from a Lean repository, performs GRPO training based on a user-defined reward function, and conducts proof search to prove sorry theorems in the traced repository.
\begin{lstlisting}[language=Python]
import torch

from lean_dojo_v2.agent.hf_agent import HFAgent
from lean_dojo_v2.trainer.grpo_trainer import GRPOTrainer


def reward_func(completions, **kwargs):
    return torch.tensor([1.0] * len(completions))


url = "https://github.com/durant42040/lean4-example",
commit = "b14fef0ceca29a65bc3122bf730406b33c7effe5"

trainer = GRPOTrainer(
    model_name="deepseek-ai/DeepSeek-Prover-V2-7B",
    output_dir="outputs-deepseek",
    reward_func=reward_func,
    epochs_per_repo=1,
    batch_size=8,
    lr=2e-5,
)

agent = HFAgent(trainer=trainer)
agent.setup_github_repository(url=url, commit=commit)
agent.train()
agent.prove()
\end{lstlisting}

\subsection*{Example 3: Theorem proving with HuggingFace API}
The following code extract sorry theorems from the Lean repository and, without training, performs whole-proof generation using the \texttt{DeepSeek-Prover-V2-671B} model provided by the HuggingFace API.
\begin{lstlisting}[language=Python]
from lean_dojo_v2.agent import ExternalAgent

url = "https://github.com/durant42040/lean4-example"
commit = "b14fef0ceca29a65bc3122bf730406b33c7effe5"

agent = ExternalAgent(model_name="deepseek-ai/DeepSeek-Prover-V2-671B:novita")
agent.setup_github_repository(url=url, commit=commit)
agent.prove(whole_proof=True)
\end{lstlisting}
\end{document}